\documentclass{scrartcl}

\usepackage [utf8]{inputenc}
\usepackage{lmodern}
\usepackage{graphicx}
\usepackage{amsmath,amssymb}
\usepackage{booktabs}
\usepackage{graphics}

\renewcommand{\vec}[1]{\mathbf{#1}}
\newcommand{\mat}[1]{\mathbf{#1}}
\newcommand{\fp}{\text{fp}}
\newcommand{\TP}{\text{TP}}
\newcommand{\Sp}{\text{Sp}}
\newcommand{\FP}{\text{FP}}
\newcommand{\eps}{\varepsilon}
\newcommand{\R}{\mathbb{R}}
\newcommand{\qi}{\mathbf{i}}
\newcommand{\qj}{\mathbf{j}}
\newcommand{\qk}{\mathbf{k}}
\newcommand{\cj}[1]{\overline{#1}}
\newcommand{\quadric}[1]{\mathcal{#1}}
\newcommand{\SQ}{\quadric{S}}
\newcommand{\SE}[1][3]{\mathrm{SE}(#1)}

\title{Optimal Synthesis of Overconstrained 6R Linkages by Curve Evolution}
\author{Tudor-Dan Rad and Hans-Peter Schröcker}

\begin{document}

\maketitle

\begin{abstract}
  The paper presents an optimal synthesis of overconstrained linkages, based on the factorization of rational curves (representing one parametric motions) contained in Study's quadric. The group of Euclidean displacements is embedded in a affine space where a metric between motions based on the homogeneous mass distribution of the end effector is used to evolve the curves such that they are fitted to a set of target poses. The metric will measure the distance (in Euclidean sense) between the two resulting vectors of the feature points displaced by the two motions. The evolution is driven by the normal velocity of the curve projected in the direction of the target points. In the end we present an example for the optimal synthesis of an overconstrained $6R$ linkage by choosing a set of target poses and explaining in steps how this approach is implemented.
 \end{abstract}

\section{Introduction}

A linkage is a mechanism which generates a complex motion. The synthesis of a linkage means determining its geometric structure such that it generates a predetermined motion or trajectory and satisfies some structural restrictions. Fulfilling the previous requirements puts a lot of limitations on the linkage. This gave rise to optimal synthesis which aims at approximating these requirements. Some of the optimization techniques used in optimal synthesis of linkages are: interior-point methods \cite{journal zhang IPM}, Gauss constraint methods \cite{journal paradis GCM} , genetic algorithms \cite{journal Cabrera GA} and evolution \cite{contrib EB}.

The paper also takes an evolutionary approach to synthesis. The novelty consists in using the factorization of motion polynomials \cite{journal factorization RJ} synthesis process. We demonstrate that it is particularly well-suited for evolution techniques because it allows to construct (overconstrained) linkages directly from a given approximated rational motion.

The factorization of motion polynomials is a process that generates a linkage which performs a one parametric motion (the functions of the joint angles share the same parameter). This motion must be defined by a rational curve in the kinematic image space. The motion curve is constructed by starting from a set of target points in this space that resemble the poses needed to be achieved by the linkage. Using curve evolution methods \cite{journal evolution curve}, the initial motion curve will converge and approximate the specified poses.

For the evolution process to work for such curves the group of Euclidean displacements $\SE$ is embedded in $12$-dimensional affine space $\mathbb{R}^{12}$. A \emph{metric} between two motions is used by equipping the end effector of the linkage with a homogeneous mass distribution or a set of ``feature points'' whose barycenter is the tool center point (TCP) \cite{journal motion metric}. The metric will measure the distance (in Euclidean sense) between the two resulting vectors of the feature points displaced by the two motions.

The paper is structured as follows Section~\ref{sec:preliminaries} explains the Euclidean metric in the affine space $\mathbb{R}^{12}$ and offers a quick glimpse into motion factorization and overconstrained linkage construction, Section~\ref{sec:evolution} presents the evolutionary design of the motion curve. Section~\ref{sec:example} follows up with an example and in Section~\ref{sec:conclusions} some conclusions are drawn.

\section{Preliminaries}
\label{sec:preliminaries}

The group of special Euclidean displacements $\SE$ represents rigid body displacements and is used to map a point $\vec{p}$ to a new position $\vec{p}'$ in Euclidean three-dimensional space:
\begin{equation}
  \gamma\colon \mathbb{R}^3 \rightarrow \mathbb{R}^3,
  \quad
  \vec{p}' = \mat{A}\vec{p}+\vec{a}.
\end{equation} 
The matrix $\mat{A}$ is a $3\times3$ special orthogonal matrix representing an element of the rotation group $\mathrm{SO}(3)$ and the vector $\vec{a}$ is a translation vector. Because displacements incorporate multiple distance concepts defining a metric between them can be problematic. In the past the concept was addressed for example by \cite{contrib E metric} but due to the nature of motion design used in this paper we have chosen the method proven in motion design by \cite{journal OB metric 2}. This approach embeds $\SE$ in a $12$-dimensional affine space by mapping the entries of $\mat{A}$ and $\vec{a}$ to a $12$-dimensional vector. In an object oriented metric the gripper is given by a set of ``feature points'' $\fp_i$ and the Euclidean metric is defined by the inner product $\langle \alpha, \beta \rangle := \sum_i \langle \alpha(\fp_i), \beta(\fp_i) \rangle$ for any $\alpha$, $\beta \in \SE$. The corresponding squared distance is $\Vert \alpha-\beta \Vert^2 = \langle \alpha-\beta, \alpha-\beta \rangle$. It is well-known \cite{journal motion metric} that this metric only depends on the barycenter and the inertia tensor of the set of feature points and is capable of representing more general mass distributions in a computationally simple way.

Motion factorization is a method developed by Hegedüs, Schicho and Schröcker in
\cite{journal factorization RJ} and can be used to synthesize linkages with one
degree of freedom joints whose end link motion is defined by a rational curve on
Study's quadric. By combining multiple factorizations overconstrained linkages
can be constructed as was demonstrated in \cite{journal factorization RJ}.

For further understanding of the synthesis process a quick introduction to the
kinematic image space and Study's quadric is necessary. Study's kinematic
mapping maps the group $\SE$ to a quadric in seven dimensional projective space
$ \mathbb{P}^7$ with the equation $x_0 y_0+x_1 y_1+x_2 y_2+x_3 y_3=0$ called
Study's quadric and denoted by $\SQ$. A more detailed explanation is given by
Husty and Schröcker in \cite{Study mapping}. The points of $\mathbb{P}^7$ are
represented by the skew ring of dual quaternions $\mathbb{DH}$, denoted as
$q=x_0+x_1\qi+x_2\qj+x_3\qk + \eps(y_0+y_1\qi+y_2\qj+y_3\qk) $ with the
multiplication properties:
\begin{equation*}
  \eps^2 = 0,\quad
  \qi^2 = \qj^2 = \qk^2 = \qi\qj\qk = -1,\quad
  \eps\qi = \qi\eps,\quad
  \eps\qj = \qj\eps,\quad
  \eps\qk = \qk\eps.\quad
\end{equation*}
The conjugate of a dual quaternion is given by replacing $\qi$, $\qj$ and $\qk$
with $-\qi$, $-\qj$, and $-\qk$, respectively. A dual quaternion on $\SQ$ is
characterized by $q\cj{q} \in \R$. 

The motion factorization algorithm of \cite{journal factorization RJ} starts
with a rational curve of degree $n$ on $\SQ$ given by the polynomial
$P(t)=c_0t^n+c_1t^{n-1}+ \cdots+c_n $ where $c_\ell \in \mathbb{DH}$ and
$P\cj{P} \in \R[t]$. Generically (only generic cases are relevant for evolution
based synthesis), it can be factored as $P(t)=(t-h_1) \cdot \ldots \cdot
(t-h_n)$. The linear factors $t-h_\ell$ are computed by polynomial division over
the dual quaternions using the quadratic irreducible factors $M_i$ of
$P\cj{P}=M_1M_2 \cdot \ldots \cdot M_n$ one at a time in the following manner:
By polynomial division, polynomials $P_{n-1}$, $R$ are attained with $P_n=P_{n-1}M_i+R$
and $R = r_1t+r_2$. In \cite{journal factorization RJ} it was proven that the
unique dual quaternion zero $h_n = -r_1^{-1}r_2$ of $R$ gives the rightmost
factor $t-h_n$ in a possible factorization of $P_n$. To obtain the remaining
linear factors, another quadratic factor $M_k$ is chosen and the process is
repeated with $P_{n-1}$ instead of~$P_n$.

Each of the $n$ linear factors represent a revolute displacement around an axis and by consecutive multiplication to the right they form a linkage whose leftmost factor is the fixed joint and the rightmost factor is the distal joint. There are, in general, $n!$ different possibilitys for the selection order of the $M_\ell$'s. This leads to the synthesis of $n!$ different open chains that perform the same motion. As it was shown in \cite{journal factorization RJ} an overconstrained linkage can be constructed by combining multiple kinematic chains to form a closed structure.

\section{Curve evolution on Study's quadric}
\label{sec:evolution}

Curve evolution is a widely used procedure in image processing and design and of
late is also used in motion generation \cite{journal OB metric}. Our
evolutionary approach is based on curve fitting to a set of data points driven
by the normal velocity of the curve in the direction of the target points
\cite{Curve Evolution}. By mapping the desired poses to $\SQ$ is obtained the set of
target points $\TP_m$ which need to be approximated by a rational curve $C$ also
contained in $\SQ$, that is, satisfying $C\cj{C} \in \R[t]$. The
validity of this condition is ensured throughout the evolution process by writing $C$ in
factorized form $C = (t-h_1) \cdots (t-h_n)$ where each linear factor represents
a rotation about an axis in space. The linear factors are defined in
\eqref{eq:1} where the Plücker coordinates of the revolute axes are $(\vec{d_i} , \vec{m_i})$:
\begin{equation}
  \label{eq:1}
  t-h_i = \frac{t-h_{0i} - \vec{d_i} - \eps\vec{m_i}}{\Vert \vec{d_i} \Vert}.
\end{equation}

By fluctuating the shape parameters $\Sp_1, \ldots ,\Sp_k$ (coefficients of $C$)
in time a family of curves $C_k$ is obtained such that the target points $\TP_1,
\ldots,\TP_m$ are optimally approximated. The moving velocity of a curve point
$C(t_j)$ is given by the amount of change in time of the shape parameters
$\dot{\Sp_k}$:
\begin{equation}\label{eq:2}
v_{Cj} = \sum \limits_{l=1}^k \dfrac{\partial C(t_j)}{\partial Sp_l} \dot{Sp_l}
\end{equation}
We are interested in moving the points on $C(t)$ which are closest to the target
points. These points are computed as the foot-normals between the $\TP_m$ and
$C(t)$ using the Euclidean structure given by the inner product defined in Section~\ref{sec:preliminaries}:
\begin{equation}
  \label{eq:3}
  \langle \TP_m-C(t) , C'(t) \rangle = 0 \leadsto \{ t_{m1}, \ldots ,t_{ml}\},
\end{equation}
\begin{equation}\label{eq:4}
t_m = \arg\min (\Vert \TP_m - C(t_{mi}) \Vert^2 : i \in \{1,\ldots, l\}).
\end{equation}
Note that the involved computations essentially boils down to finding the zeros
of a univariate polynomials because the motion is given by a polynomial $C$.
This is one of the advantages inherent to our approach.

The foot-points $\FP_m = C(t_m)$ are computed using relations \eqref{eq:3} and
\eqref{eq:4} and so the ideal velocity vector $\vec{\hat{d}}$ of the foot-points
should be $\TP_m-\FP_m$. Comparing coefficients of both vectors in an
orthonormal basis (with respect to the given scalar product) and using
\eqref{eq:2} results in an overconstrained system of linear equations for
$\dot\Sp_l$ that can be solved in least square sense. The new shape parameters
of the curve are computed as: $\Sp_l = \Sp_l + \lambda \dot{\Sp_l}$ where
$\lambda$ is a scaling parameter used such that the curve doesn't overshoot. In
time as the distance between the curve $C(t)$ and the target points $\TP_m$
decreases, the system will converge to a local minimum.

\section{Numeric Example}
\label{sec:example}

For the example, a set of $11$ target poses is chosen as shown in
Table~\ref{table:param}. Without loss of generality, the first pose is the
identity. A cubic curve $ C(t) = (t-\vec{x})(t-\vec{y})(t-\vec{z}) $ is chosen
to approximate the target poses. We limit ourselves to polynomials of degree
three because the end goal of the example is to construct a $6$R overconstrained
linkage. More explicitly, the linear factor $t-\vec{x}$ is of the shape
\begin{equation}\label{eq:6}
  \frac{t-x_0+x_1 \qi+x_2 \qj+x_3 \qk-\eps ((x_2 x_7-x_3 x_6) \qi+(x_3 x_5-x_1 x_7) \qj+(x_1 x_6-x_2 x_5) \qk)}{\sqrt{x_1^2+x_2^2+x_3^2}},
\end{equation}
and similar for $t-\vec{y}$ and $t-\vec{z}$, resulting in a total of $21$ shape
parameters. The special shape of \eqref{eq:6} is crucial. It ensures validity of
the Study condition for each factor and hence also for $C(t)$ \emph{throughout
  the whole evolution process.} The linear factors are normalized to avoid
numeric fluctuation of $C(t)$ without any geometric change.

\begin{table}
  \caption{Target Poses}
  \label{table:param}
  \centering
  \begin{tabular}{p{0.7cm}p{9cm}}
    \midrule\noalign{\smallskip}
    \textbf{TP} & \emph{Study Parameters} \\
    \hline\noalign{\smallskip}
    1 & 1 
        \vspace{1mm}\\
    2 & $\qi (-( \frac{337}{7}) \eps- \frac{79}{968})+\qj ((\frac{22}{379}) \eps+ \frac{3}{5066})+\qk (( \frac{67}{161} ) \eps+ \frac{55}{509} )-( \frac{108}{53} ) \eps+ \frac{79}{41}$ 
        \vspace{1mm}\\
    3 & $\qi (-(\frac{2533}{19}) \eps- \frac{73}{328})+\qj ((\frac{92}{105}) \eps+ \frac{8}{3103})+\qk ((\frac{46}{89}) \eps+ \frac{41}{128})-(\frac{313}{29}) \eps+ \frac{119}{44}$ 
        \vspace{1mm}\\
    4 & $ \qi (-(\frac{949}{4}) \eps-\frac{30}{77})+\qj ((\frac{163}{43}) \eps+\frac{7}{815})+\qk (-(\frac{51}{128}) \eps+\frac{153}{245})-(\frac{528}{19}) \eps+\frac{95}{29} $
        \vspace{1mm}\\
    5 & $ \qi (-(\frac{1696}{5}) \eps-\frac{110}{201})+\qj ( (\frac{247}{21}) \eps+ \frac{11}{456})+\qk (-(\frac{89}{24}) \eps+\frac{906}{907})-(\frac{660}{13}) \eps+\frac{165}{46} $
        \vspace{1mm}\\
    6 & $ \qi (-(\frac{2939}{7}) \eps-\frac{79}{119})+\qj ((\frac{1279}{46}) \eps+\frac{19}{358})+\qk (-(\frac{1449}{145}) \eps+\frac{100}{71})-(\frac{816}{11}) \eps+\frac{18}{5} $
        \vspace{1mm}\\
    7 & $ \qi (-461 \eps-\frac{487}{682})+\qj ((\frac{575}{11}) \eps+\frac{10}{101})+\qk (-(\frac{557}{32}) \eps+\frac{135}{76})-(\frac{185}{2}) \eps+\frac{257}{78} $
        \vspace{1mm}\\
    8 & $ \qi (-(\frac{1364}{3}) \eps-\frac{115}{167})+\qj ((\frac{864}{11}) \eps+\frac{31}{212})+\qk (-(\frac{347}{14}) \eps+\frac{291}{145})-(\frac{705}{7}) \eps+\frac{191}{70} $
        \vspace{1mm}\\
    9 & $ \qi (-(\frac{1226}{3}) \eps-\frac{44}{73})+\qj ((\frac{957}{10}) \eps+\frac{181}{1018})+\qk (-(\frac{817}{27}) \eps+\frac{75}{37})-(\frac{694}{7}) \eps+\frac{390}{191} $
        \vspace{1mm}\\
    10 & $ \qi (-(\frac{1713}{5}) \eps-\frac{85}{173})+\qj ((\frac{1921}{20}) \eps+\frac{20}{109})+\qk (-(\frac{713}{22}) \eps+\frac{106}{57})-(\frac{715}{8}) \eps+\frac{38}{27} $
        \vspace{1mm}\\
    11 & $ \qi (-(\frac{8831}{32}) \eps-\frac{77}{201})+\qj ((\frac{248}{3}) \eps+\frac{52}{311})+\qk (-(\frac{458}{15}) \eps+\frac{73}{46})-(\frac{1476}{19}) \eps+\frac{109}{119} $
        \vspace{1mm}\\
    \midrule\noalign{\smallskip}
  \end{tabular}
\end{table}

A suitable initial guess for the shape parameter can found by interpolating four
poses \cite{SQ cubic interp} or, as we did in our example, by assigning random
values to the shape parameters. Several attempts might be necessary in order to
ensure good convergence. Once the evolution runs smoothly little effect on the
local minimum has been observed. In the first iterations, the scaling factor
$\lambda$ needs to be small enough in order to compensate for large amount of
changes $\dot{\Sp_l}$ in the shape parameters. For the evolution to have a good
flow, we found $\lambda := \max\{10\Vert \dot{\Sp_l} \Vert^{-1}_\infty,1\}$ to
be a good choice. With this initial setup we arrive at the computation of the
foot points on $C(t)$ as described in the previous section. From relation
\eqref{eq:3} we obtain an equation of degree at most $10$. Its zeros are found
numerically and we can use \eqref{eq:4} to find the parameter value of the
closest point. Here, we also impose some constraints on the foot point
computation in order to ensure that the poses are visited in successive order.
To achieve this, after a provisional curve is evolved, two target points are
chosen which are approximated best. Next interval constraints are applied to the
remaining foot points such that their respective parameter values $t_m$ are in
successive order. If the computed values of foot points do not fit the
constraint interval, the boundary point with the minimal distance is chosen.

\begin{figure}
 \includegraphics[width=0.48\linewidth,trim=300 0 0 0,clip]{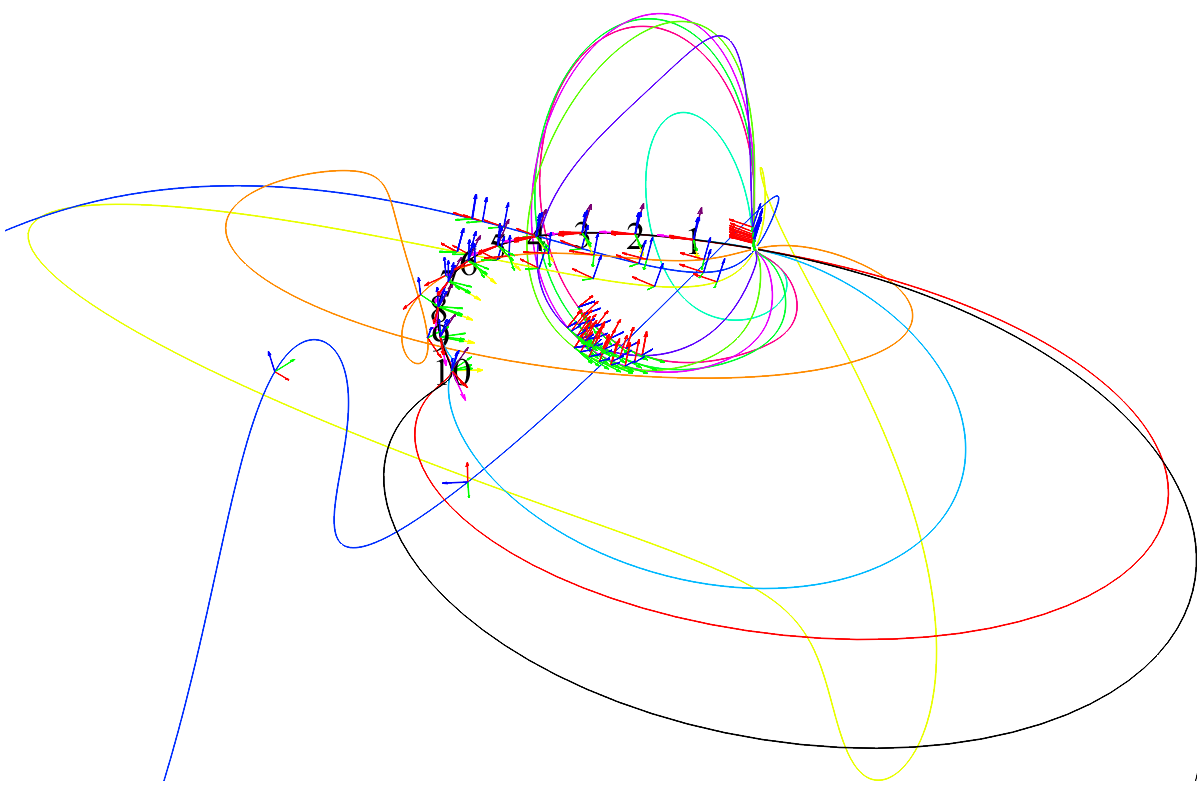}\hfill
  \includegraphics[width=0.48\linewidth,trim=70 0 150 0,clip]{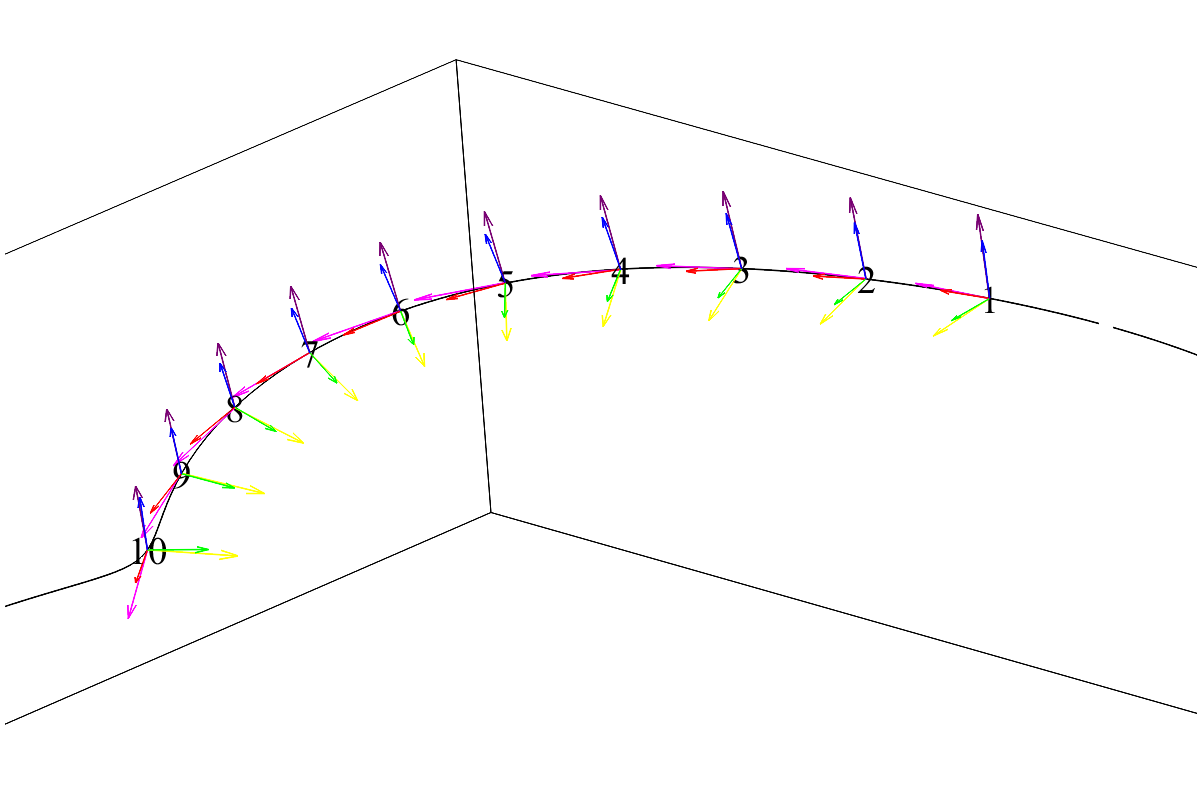}
 \caption{TCP trajectory and orientation during the evolution process}
 \label{fig:1}
\end{figure}

The evolution rule consists of comparing coordinates of the curve velocity
\eqref{eq:2} with respect to some orthonormal basis of $\R^{12}$ with the
coordinates of the difference vector from foot point to target point. This
produces an overconstrained system which is solved in least square sense. Hence
the shape parameters $\Sp_{pl}$ approximately change with the corresponding
amount and the process is repeated again starting with the foot point
computation. The final results are presented in Table \ref{table:rezulst} with a
three decimal digit precision. The evolution process itself is visualized in
Figure~\ref{fig:1}. The target poses are labelled from 1 to 10, the angles and
distances to the respective target poses are given in
Table~\ref{table:variation}. It can be seen that the distances are quite good
while the orientation seems to be hard to match. The reasons for this are under
investigation. We conjecture an inappropriate distribution of feature points.

\begin{table}
  \caption{Final Shape Parameters}
  \label{table:rezulst}
  \centering
  \begin{tabular}{ccccccc} 
    \midrule\noalign{\smallskip}
    $x_0$ & $x_1$ & $x_2$ & $x_3$ & $x_5$ & $x_6$ & $x_7$ \\
    $5.822$ & $-0.213$ & $0.2$ & $-0.337$ & $-329.055$ & $82.644$ & $-100.544$ \\
    \hline\noalign{\smallskip}
    $y_0$ & $y_1$ & $y_2$ & $y_3$ & $y_5$ & $y_6$ & $y_7$ \\
    $6.084$ & $0.051$ & $-0.244$ & $0.287$ & $-8.987$ & $-749.392$ & $937.288$ \\
    \hline\noalign{\smallskip}
    $z_0$ & $z_1$ & $z_2$ & $z_3$ & $z_5$ & $z_6$ & $z_7$ \\
    $4.926$ & $0.061$ & $0.181$ & $0.384$ & $-99.5$ & $-423.666$ & $34.386$ \\
    \midrule\noalign{\smallskip}
  \end{tabular}
\end{table}

\begin{table}
  \caption{Variation of angle (in radians) and distance}
  \label{table:variation}
  \centering
  \begin{tabular}{c|c @{\hskip 0.2cm} c @{\hskip 0.2cm} c @{\hskip 0.2cm} c @{\hskip 0.2cm} c @{\hskip 0.2cm} c @{\hskip 0.2cm} c @{\hskip 0.2cm} c @{\hskip 0.2cm} c @{\hskip 0.2cm} c} 
  
    \midrule\noalign{\smallskip}
    \textbf{TP \textsubscript{s}} & \textbf{TP\textsubscript{1}} & \textbf{TP\textsubscript{2}} & \textbf{TP\textsubscript{3}} & \textbf{TP\textsubscript{4}} & \textbf{TP\textsubscript{5}} & \textbf{TP\textsubscript{6}} & \textbf{TP\textsubscript{7}} & \textbf{TP\textsubscript{8}} & \textbf{TP\textsubscript{9}} & \textbf{TP\textsubscript{10}} \\
    \hline\noalign{\smallskip}  
  
    $\phi$ & $0.061$ & $0.102$ & $0.125$ & $0.139$ & $0.124$ & $0.106$ & $0.08$ & $0.061$ & $0.151$ & $0.421$ \\
    
    \hline\noalign{\smallskip}
    \textbf{t} & $0.922$ & $1.488$ & $1.38$ & $1.182$ & $1.522$ & $2.689$ & $4.914$ & $8.26$ & $3.736$ & $4.336$ \\ 

    \midrule\noalign{\smallskip}
  \end{tabular}
\end{table}

\begin{figure}
  \centering
  \begin{minipage}[b]{0.45\linewidth}
    \centering
    \includegraphics[width=1\linewidth]{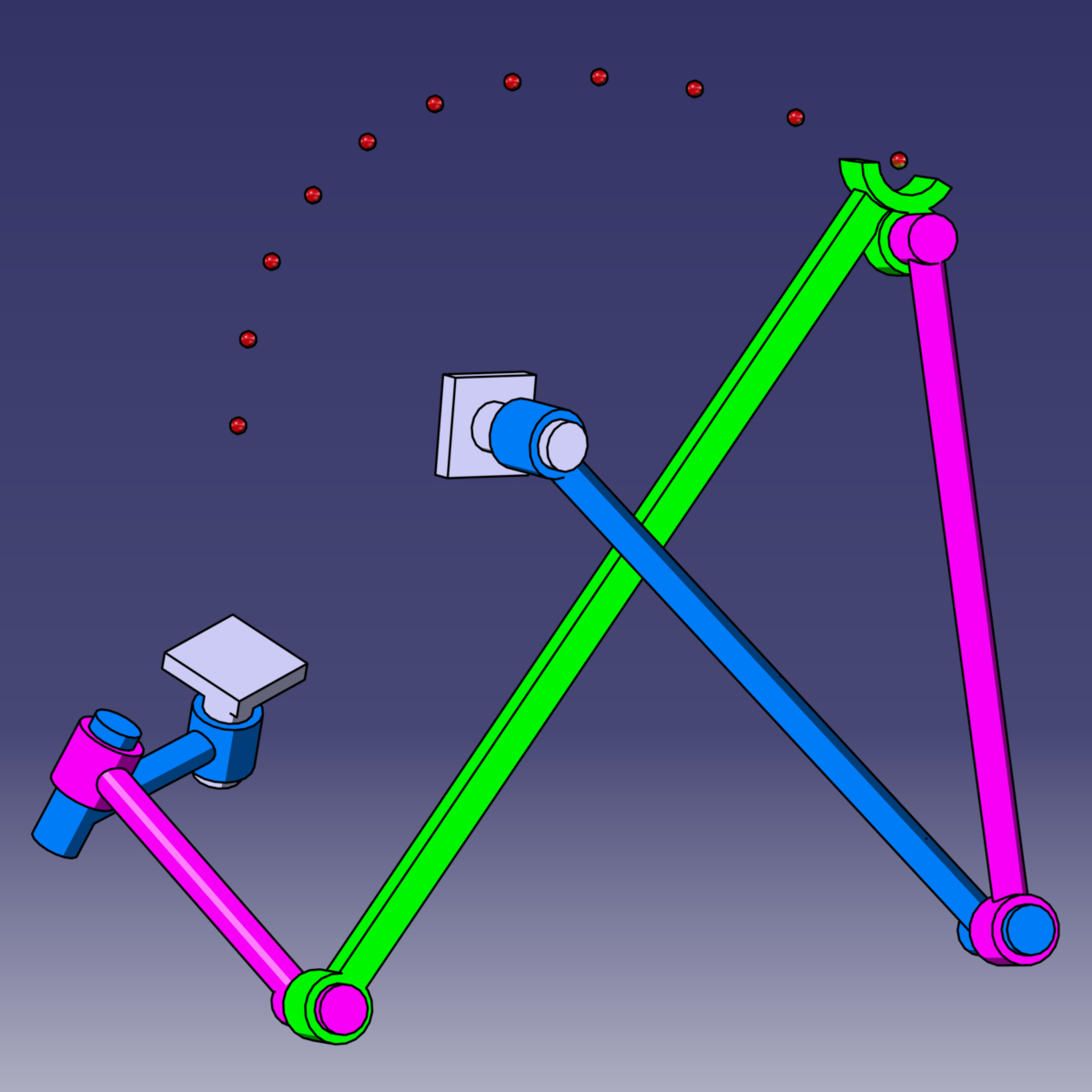} 
  \end{minipage}
  \begin{minipage}[b]{0.45\linewidth}
    \centering
    \includegraphics[width=1\linewidth]{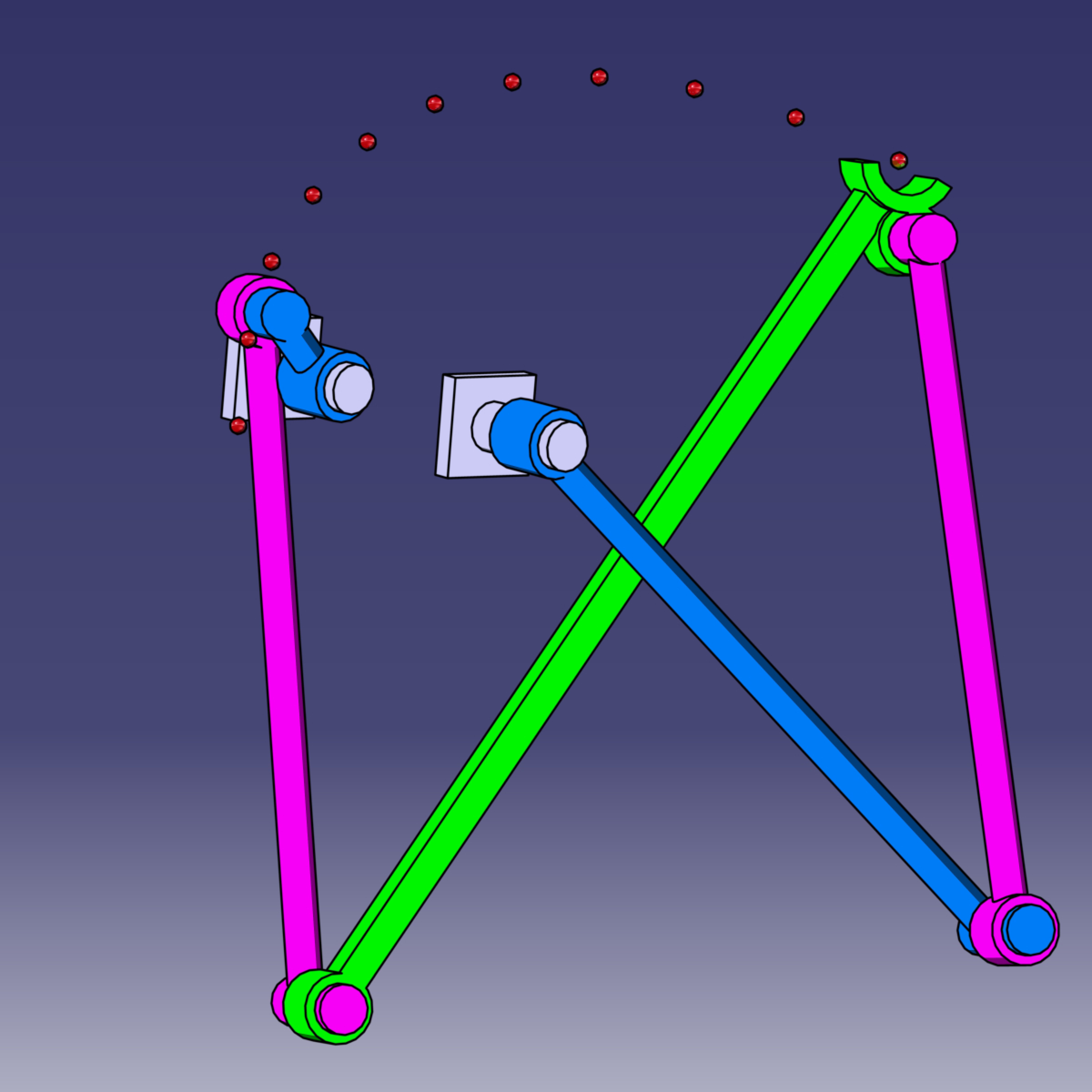} 
  \end{minipage} 
  \begin{minipage}[b]{0.45\linewidth}
    \centering
    \includegraphics[width=1\linewidth]{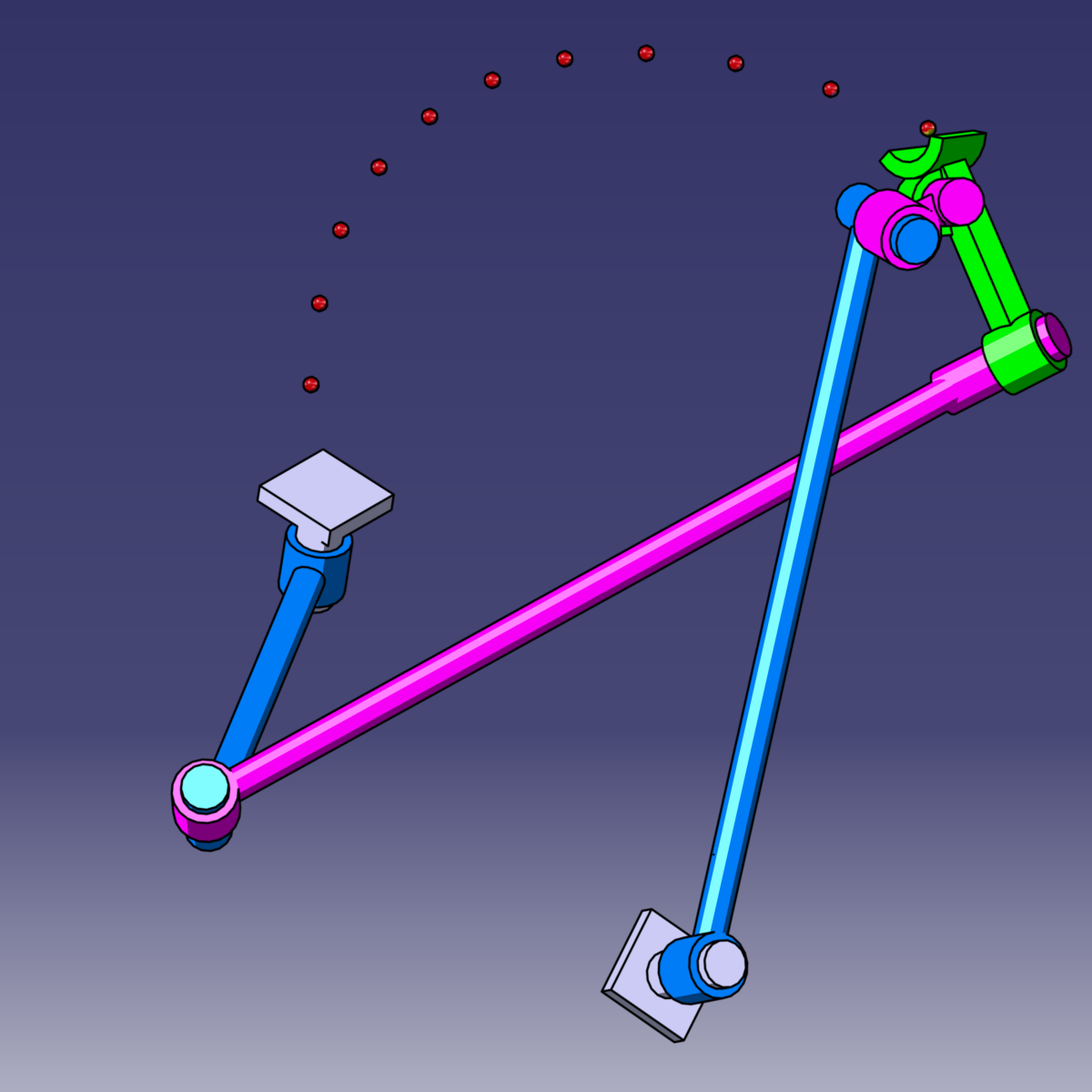} 
  \end{minipage}
  \begin{minipage}[b]{0.45\linewidth}
    \centering
    \includegraphics[width=1\linewidth]{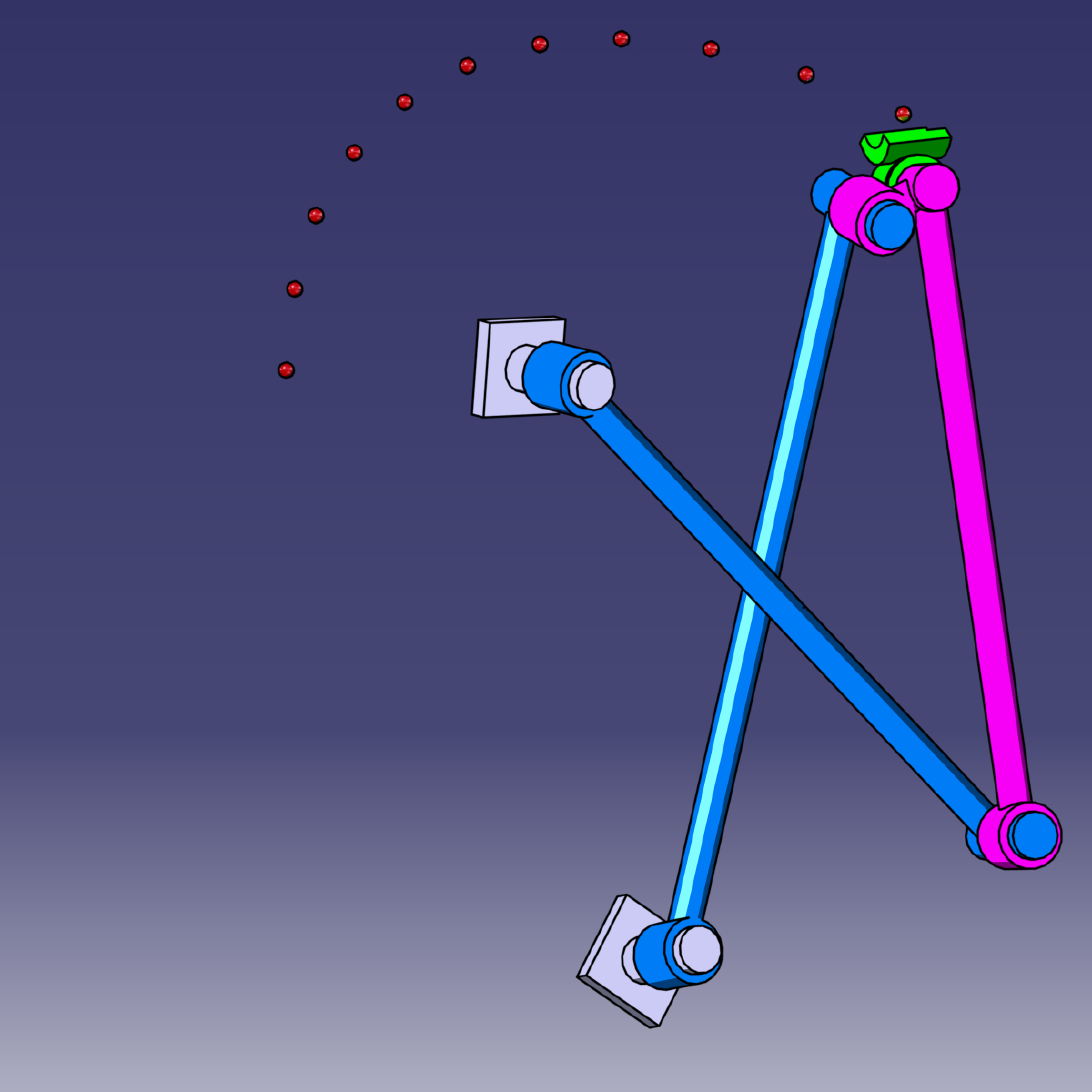} 
  \end{minipage} 
  \caption{Four different 6R linkages obtained}
  \label{fig:2} 
\end{figure}

After the motion curve $C(t) = (t-\vec{x})(t-\vec{y})(t-\vec{z})$ is obtained we can
start the synthesis of the overconstrained 6R linkage using motion factorization
\cite{journal factorization RJ} as explained in Section~\ref{sec:preliminaries}.
First the quadratic factors $M_i$ are computed by multiplying the curve with
it's quaternion conjugate:
\begin{equation}
  \label{eq:7}
  C\cj{C} = (t^2-12.165t+37.143)(t^2-11.648t+34.116)(t^2-9.853t+24.456)
\end{equation} 
By selecting the first quadratic factor from \eqref{eq:7} polynomial
division (a variant of Euclid's algorithm taking into account the non-commutativity of quaternion multiplication) is used to divide $C(t)$ and single out the remainder
\begin{equation}
  \label{eq:10}
\begin{aligned}
  &(-59.057 \qi \eps-0.191 \qi-9.9 \qj \eps+0.036 \qj-13.531 \qk \eps+ 0.134 \qk-16.841 \eps+ 0.352) t+ \\
  &+347.317 \qi \eps+1.143 \qi+70.062 \qj \eps- 0.222 \qj+80.891 \qk \eps-0.667 \qk+94.976 \eps-2.208   
\end{aligned}
\end{equation}
The constant term $h_{13}$ in the rightmost factor is computed as a unique root of
this linear remainder polynomial:
\begin{equation}
  h_{13} = \qi (30.463 \eps+0.135)+\qj (16.643 \eps+0.135)+ \qk (19.361 \eps-0.329)+6.084
\end{equation}
After the first root is computed $C(t)$, is divided by $t-h_{13}$ and the
process is iterated with the quotient and with one of the remaining quadratic
factors from \eqref{eq:7}. After the second root is computed the quotient will
be the last linear factor. All the possible combinations in which the quadratic
factor can be chosen will produce six different open 3R kinematic chains.
Suitable combinations \cite{journal factorization RJ} then give overconstrained
6R linkages. Four examples are depicted in Figure~\ref{fig:2}
 
\section{Conclusions}
\label{sec:conclusions}

We used properties of the factorized representation of rational motions to
set-up an evolution process for optimal design of corresponding linkages. The
evolution gives an open kinematic chain that, if desired, can be combined with
other chains obtained from different factorizations to produce overconstrained
linkages. From a mechanical point of view, overconstrained linkages are robust,
need minimal control elements and they are ideal for repetitive motions in an
interval. In Section~\ref{sec:example} we illustrated this process for an
overconstrained 6R linkage. So far, position matching is good while matching
orientations should be improved.

The construction relies on the factorized representation which helps to ensure
validity of the Study condition throughout the evolution and automatically
relates the rational motion to kinematic chains. Moreover, rationality allows
efficient and stable computation of footpoints which is a crucial part in any
evolution based mechanism synthesis.

  The research was supported by the Austrian Science Fund (FWF): P\;26607
  (Algebraic Methods in Kinematics: Motion Factorisation and Bond Theory).

\end{document}